\pdfoutput=1

\documentclass[11pt]{article}

\usepackage[final]{acl}

\usepackage{times}
\usepackage{latexsym}

\usepackage[T1]{fontenc}

\usepackage[utf8]{inputenc}

\usepackage{microtype}

\usepackage{inconsolata}

\usepackage{graphicx}
\usepackage{graphicx}
\usepackage{times}
\usepackage{latexsym}
\usepackage{amsmath} 
\usepackage{amssymb} 
\usepackage{algorithm}
\usepackage{algpseudocode}
\usepackage{graphicx} 
\usepackage{tabularx}
\usepackage{booktabs}
\usepackage{multirow}
\usepackage{booktabs}
\usepackage{wrapfig}
\usepackage{caption}
\usepackage{enumitem}
\usepackage{tablefootnote}
\usepackage{array}
\usepackage{setspace}
\usepackage{CJKutf8} 
\usepackage{graphicx}
\usepackage{float}
\usepackage{subfigure}
\usepackage[T1]{fontenc}

\usepackage{graphicx}

\newcommand\blfootnote[1]{%
\begingroup
\renewcommand\thefootnote{}\footnote{#1}%
\addtocounter{footnote}{-1}%
\endgroup
}
\title{ MoR: Mixture of Ranks for Low-Rank Adaptation Tuning}

\author{
    Chuanyu Tang$^{1,2,*}$,
    ~Yilong Chen$^{1,2,*}$,
    ~Zhenyu Zhang$^{3}$,
    ~Junyuan Shang$^{3}$,\\
    ~\textbf{Wenyuan Zhang}$^{1,2}$,
    ~\textbf{Yong Huang}$^{1,2}$,
    ~\textbf{Tingwen Liu}$^{1,2\dagger}$ \\ 
    \normalsize $^1$ Institute of Information Engineering, Chinese Academy of Sciences\\
    \normalsize $^2$ School of Cyber Security, University of Chinese Academy of Sciences\\
    \normalsize $^3$ Baidu Inc.\\
     \{\texttt{tangchuanyu, chenyilong, zhangwenyuan, huangyong, liutingwen\}@iie.ac.cn} \\
     \{\texttt{zhangzhenyu07, shangjunyuan\}@baidu.com} \\
}

\begin{document}
\maketitle
\begin{abstract}
Low-Rank Adaptation (LoRA) drives research to align its performance with full fine-tuning.
However, significant challenges remain: (1) Simply increasing the rank size of LoRA does not effectively capture high-rank information, which leads to a performance bottleneck. 
(2) MoE-style LoRA methods substantially increase parameters and inference latency, contradicting the goals of efficient fine-tuning and ease of application. To address these challenges, we introduce Mixture of Ranks (MoR), which learns rank-specific information for different tasks based on input and efficiently integrates multi-rank information. We firstly propose a new framework that equates the integration of multiple LoRAs to expanding the rank of LoRA. Moreover, we hypothesize that low-rank LoRA already captures sufficient intrinsic information, and MoR can derive high-rank information through mathematical transformations of the low-rank components. Thus, MoR can reduces the learning difficulty of LoRA and enhances its multi-task capabilities. MoR achieves impressive results, with MoR delivering a 1.31\% performance improvement while using only 93.93\% of the parameters compared to baseline methods. 
\end{abstract}

\blfootnote{$^*$ Equal contribution. $^\dagger$ Corresponding author.}
\section{Introduction}

Large language models (LLMs)~\cite{GPT-4,claude} based on the Transformer architecture have made significant breakthroughs in various natural language tasks due to their powerful understanding and generation capabilities. When fine-tuned on downstream datasets, these models demonstrate outstanding performance across various tasks~\cite{touvron2023llama,jiang2023mistral}. However, their ever-increasing size requires substantial computational resources for comprehensive fine-tuning. Parameter-efficient fine-tuning~\cite{ding2023parameter}, particularly Low-Rank Adaptation (LoRA)~\cite{hu2021lora}, is widely used by freezing pre-trained model weights and injecting trainable rank-decomposition matrices. In multi-task scenarios, recent research has begun integrating multiple LoRAs~\cite{liu2023moelora,dou-etal-2024-loramoe} or exploring increasing the rank of LoRA~\cite{wang2024prolora} to enhance performance or prevent catastrophic forgetting.

\begin{figure}[t]  
\centering  
\includegraphics[width=7cm]{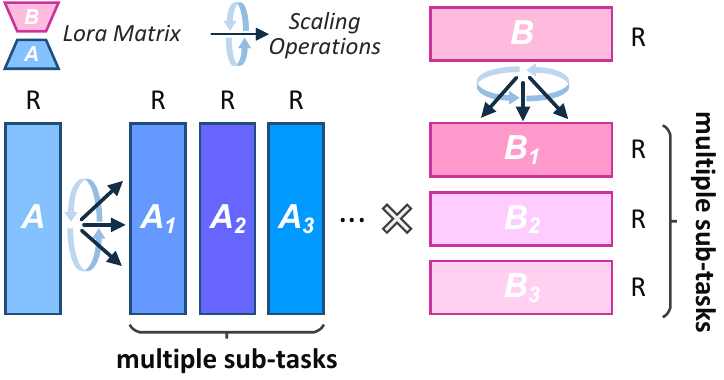}
\caption{Integrating multiple LoRAs in MoR can be regarded as increasing the rank of LoRA. MoR learns a shared parameter, which is then mapped to each subtask space through transformation. MoR focuses the learning goal on a small amount of effective information, greatly reducing the learning cost while increasing rank.}
\label{fig:motivation}
\end{figure}

However, directly combining multiple LoRAs increases the cost of parameter-efficient fine-tuning substantially~\cite{liu2023moelora}, especially when the number of LoRAs is large or the rank is high. Increasing the number of LoRAs results in learning redundant representations, failing to capture the essential differences between tasks in multi-task settings~\cite{wang2023multilora}. On the other hand, increasing the rank of LoRA presents a performance bottleneck~\cite{hu2021lora}, as learning intrinsic task-specific representations with LoRA proves challenging~\cite{liu2024dora}. Therefore, we aim to address the following critical question:

\noindent\textit{How can we enhance the model's multi-task capabilities with high-level parameter efficiency?}

To answer this question, we first analyze current work on LoRA integration~\cite{liu2023moelora,dou-etal-2024-loramoe} and propose an interpretive framework that suggests the fundamental objective of MoE-style LoRA is to \textit{softly} increase LoRA's ranks to improve performance. Given that this objective faces the challenge of LoRA’s difficulty in capturing essential multi-task information by merely increasing the rank~\cite{hu2021lora}, we propose a new approach: \textit{learning a shared parameter subspace that represents the fundamental objectives of fine-tuning and then mapping this parameter subspace onto multiple sub-tasks through transformations}, as shown in Figure~\ref{fig:motivation}. This allows the learning process to focus on effectively capturing intrinsic information, requiring only a small number of parameter adaptations to enhance multi-task capabilities.

Based on this idea, we introduce the Mixture of Ranks (MoR) method. This approach consists of three main components: shared experts, multi-rank adaptation, and mixture learning. Inspired by parameter-sharing methods~\cite{wang2024prolora} and shared Mixture of Experts (MoE) methods~\cite{liu2024deepseek}, we learn a shared LoRA when keeping the parameter matrix frozen. To adapt to multiple task objectives, we simultaneously train multiple mapping vectors for each LoRA, which are used to learn scaled experts for different forms of the LoRA. Finally, MoR assigns inputs to different mapped LoRA experts via routing and dynamically learns the optimal combination of expert outputs based on the input and specified task objectives. This approach dramatically reduces the training cost of multi-LoRA systems and dynamically learns more concise information across tasks.

Our experiments demonstrate the effectiveness, efficiency, generalizability, and scalability of the MoR method. In empirical studies on multiple instruction-tuning datasets and evaluations on 11 downstream metrics, MoR achieved an improvement of \textit{1.31\%} compared to MoELoRA and \textit{7.4\%} compared to LoRA with equal tunable parameters. Ablation studies show that MoR consistently demonstrates significant improvement when increasing the rank size and expanding the base model size. Overall, MoR represents an outstanding practice, balancing efficiency and performance.

\section{Preliminary} \label{section:pre}




In this section, we briefly introduce the background and principles of MoE-style LoRA and demonstrate how MoE-style LoRA leverages sparse activation to construct an effective high-rank structure from low-rank LoRA modules.

\paragraph{When MoE adapted to LoRA.}
The traditional MoE structure is incorporated into model fine-tuning by employing multiple expert modules, which can be expressed as:
\begin{equation}
\begin{aligned}
     \Delta W x = \sum_{i=1}^{N}\mathcal{G}_i(x)E_i(x)
\end{aligned}
\end{equation}
where $W \in \mathbb{R}^{d_{out} \times d_{in}}$ is the pre-trained model original parameter. After converting the network into MoE structure, the entire module is replaced with multiple isomorphic or heterogeneous sub-expert network modules $E_i$. The gated routing $\mathcal{G}_i$ is used to calculate expert weights and allocate routing for different input tokens.



Considering that the MoE networks have a large number of parameters and require extensive resources to maintain model resilience and continue pre-training, it is impractical to use MoE approach for instruction supervision fine-tuning.
A feasible approach is using LoRA module to implement the MoE model. Specifically, each MoE expert $E_i$ will be replaced by LoRA matrices $A_i$ and $B_i$, and it can be expressed as $E_i = B_i A_i$. Therefore, the MoE-style LoRA network plugin module can be denoted as:
\begin{equation}
\begin{aligned}
   o &= Wx + \Delta Wx \\
     &= Wx + \sum_{i=1}^{N} \mathcal{G}_i(x) B_i A_i x
\end{aligned}
\end{equation}
where $B \in \mathbb{R}^{d \times r} $ and $A \in \mathbb{R}^{r \times h} $ denotes the LoRA matrice, where $r $ denotes the rank of the matrix, and N denotes the number of LoRA experts.

\paragraph{MoE-style LoRA plugin is a High-Rank LoRA with Sparse Activation.} We begin by considering a high-rank LoRA, where the updated matrix $\Delta W = B A $
captures information across $r $ dimensions. To decompose this high-rank structure, we introduce $n $ low-rank LoRA modules, each with a rank of $\frac{r}{n} $, such that the update matrix for the $i $-th module is given by $\Delta W_i = B_i A_i $, where $B_i \in \mathbb{R}^{d \times \frac{r}{n}} $ and $A_i \in \mathbb{R}^{\frac{r}{n} \times h}$, $B_i = B\left[:, \frac{i r}{n} : \frac{(i+1) r}{n}\right], A_i = A\left[\frac{i r}{n} : \frac{(i+1) r}{n}, :\right]$. These matrices $B_i $ and $A_i $ represent sub-blocks of the original matrices $B $ and $A $, respectively.

The original high-rank matrix $\Delta W $ can then be reconstructed by summing over the contributions of all low-rank modules as:
\begin{equation}
\Delta W = B A = \sum_{i=1}^{n} B_i A_i
\end{equation}
This decomposition shows that the high-rank matrix is, in fact, a combination of several lower-rank matrices. In the context of MoE-LoRA, this decomposition is further enhanced by introducing sparse activation. MoE activates only a subset of experts for any given input, and the output is the weighted sum of these activated experts. For $k $ activated experts, the update matrix becomes:
\begin{equation}
\Delta W_{\text{MoE}} = \sum_{i=1}^{k} \mathcal{G}_i(\cdot) B_i A_i
\end{equation}
where $\mathcal{G}_i $ represents the gating function’s weight for the $i $-th expert. By selectively activating experts, MoE effectively constructs a higher-rank representation from a subset of low-rank modules. This sparse activation ensures computational efficiency while maintaining the expressive power of a high-rank matrix.

\begin{figure*}[t]  
\centering  
\includegraphics[width=16cm]{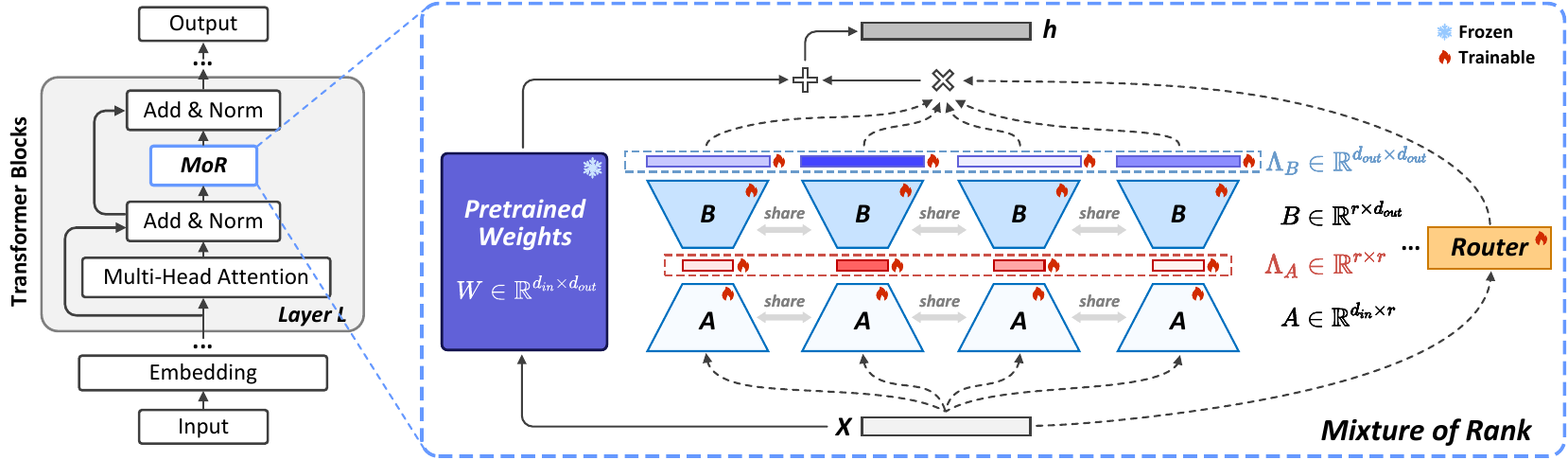}
\caption{Overview of the MoR framework, this method takes place the raw FFN module of LLMs with the MoR plugin. The LoRA matrix $A$ and $B$ are shared, and the multi-dimensional scaling vector diagonal matrix $\Lambda$ is specialized with different domain information, mapping the shared LoRA matrices into specific subspace.}
\label{fig:method}
\end{figure*}

\paragraph{Advantages and Challenges of increasing LoRA rank.} High-rank LoRA retains a more complete representation of the original model’s weight matrix $W $. From the perspective of singular value decomposition (SVD)~\cite{wall2003singular}, $W $ can be expressed as $W = U \Sigma V^T $, where $\Sigma $ contains the singular values. In LoRA with rank $r $, truncates smaller singular values, potentially losing valuable information. By increasing the rank $r $, high-rank LoRA retains more singular values, thus minimizing the information loss, which can be summarized as:
\begin{equation}
\begin{aligned}
\| \Delta W - \Delta U_r \Sigma_r V_r^T \|_F^2 \to 0 \quad \\ \text{as} \quad r \to \min(d, h)
\end{aligned}
\end{equation}
where $\|\cdot\|_F $ is the Frobenius norm. This higher fidelity approximation allows the model to capture more intricate task-specific features, improving performance in complex tasks by preserving a broader spectrum of information in the parameter space.

However, the increased rank also brings challenges. High-rank LoRA introduces more parameters, which raises the risk of overfitting, particularly when the additional capacity exceeds the task's complexity~\cite{lin2024lora}. This increased parameter space can obscure task-relevant information and make learning more difficult~\cite{zi2023deltalora}. To address this, it is possible to leverage transformations to map low-rank LoRA into a higher-rank space. By applying such transformations, we can effectively represent multi-task information in a high-rank space while maintaining the computational efficiency of low-rank models~\cite{wang2024prolora}. This approach allows us to exploit the expressiveness of high-rank LoRA without the drawbacks of increased complexity:
\begin{equation}
B'A' = \Lambda_{B} B \Lambda_{B}^T \Lambda_{A} A \Lambda_{A}^T 
\label{eq:trans}
\end{equation}
where $\Lambda$ is a transformations matrix that expands the low-rank matrix into a higher-dimensional subspace, facilitating the learning of multi-task representations while controlling over-fitting.

\section{Method}






In this section, we introduce MoR for efficient multi-task low-rank parameter fine-tuning. MoR enhances the efficiency of parameter tuning across various domains by leveraging multi-dimensional scaling transformations on the LoRA matrix. 

By transforming the LoRA matrix, MoR can capture task-specific information while preserving shared knowledge in the standard LoRA parameters. This design enhances the model's capacity to learn high-rank spatial representations, thereby improving the performance of LLMs in multi-task learning scenarios.

\subsection{Transform of LoRA} \label{section:1}
In this section, We introduce a LoRA transformation method by applying intrinsic scaling transformations to the exact subspace of the LoRA matrix.

Applying specific transformations to the LoRA matrix enables the model to capture more intricate information~\cite{renduchintala2023tied,wang2024prolora}. To emphasize directional and magnitude-specific changes in LoRA, we introduce task-related, learnable vectors, $\lambda_A$ and $\lambda_B$, to transform the LoRA matrices $A$ and $B$. We concrete the transformation operations on LoRA matrices, which is expressed in Equation~\ref{eq:trans}. In order to reduce computation, we have performed parameter absorption on $\Lambda_B^T$ and $\Lambda_A$, and adopted $\Lambda_A^T$ into router, as follows: 
\begin{equation} 
\begin{split} 
\hat{A}= \Lambda_A A , \hspace{0.2cm} \hat{B}= \Lambda_B B
\end{split} 
\end{equation} 
where $A \in \mathbb{R} ^{r\times d_{in}}$ and $B \in \mathbb{R} ^{d_{out}\times r}$ represent the LoRA matrices $A$ and $B$ respectively, $\Lambda_A \in \mathbb{R} ^{r \times r}$ denotes the scaling matrix applied to $A$, and $\Lambda_B \in \mathbb{R} ^{d_{out} \times d_{out}}$ denotes the scaling matrix applied to $B$. 

For $\Lambda_A$ and $\Lambda_B$, we have: 
\begin{equation} 
\begin{split} 
\Lambda_A = \text{diag}(\lambda_{A_0}, \lambda_{A_1}, \dots, \lambda_{A_{r-1}}) \\
\Lambda_B = \text{diag}(\lambda_{B_0}, \lambda_{B_1}, \dots, \lambda_{B_{d_{out}-1}}) 
\end{split} 
\end{equation} 
where $\text{diag}(\cdot)$ represents a diagonal matrix with learnable parameters $\lambda_A$ and $\lambda_B$ along its diagonal. These transformations allow the rank of LoRA’s matrices $A$ and $B$ to be transformed in parameter space, within a specific subspace, enabling the learning of high-rank information. After applying the above transformations, the forward propagation in LoRA training can be formulated as: 
\begin{equation} 
\begin{split} 
o &= Wx + \Delta Wx \\
&\approx Wx+ \frac{\alpha}{r} \hat{B} \hat{A} x \\
&\approx Wx + \frac{\alpha}{r} \Lambda_B B \Lambda_A A x
\end{split} 
\end{equation} 
where $\Delta W$ is the full-rank update matrix, $\alpha$ denotes the scaling factor, and $A$ and $B$ are low-rank projection matrices.

\subsection{MoR Architecture}
In the MoE-LoRA structured model, each expert consists of projection matrices $A_i$ and $B_i$, leading to significant parameter redundancy~\cite{dou-etal-2024-loramoe,liu2023moelora}. Additionally, the division of parameters across different experts can result in losing coupling information.

To alleviate these challenges, we propose a parameter-sharing method called MoR, which leverages multiple transformation vectors to scale the shared LoRA matrices $A$ and $B$ within a specific parameter subspace. A space importance adjustment router $\mathcal{G}$ is introduced to weigh the effect of different vectors, enabling efficient parameter fine-tuning across multiple task-specific subspaces. The overall structure is illustrated in Figure~\ref{fig:method}.

For a single direction $D_i$, the transformation is defined as: \begin{equation} 
D_i = \frac{\alpha}{r}\Lambda_{B_i} B \Lambda_{A_i} A \end{equation} 
where $B \in \mathbb{R}^{d_{out} \times r}$ and $A \in \mathbb{R}^{r \times d_{in}}$ represent the shared LoRA matrices, and $\Lambda_{A}$ and $\Lambda_{B}$ are the transformation matrices of the LoRA projections, described in Section~\ref{section:1}. 
$\alpha$ is the scaling hyperparameter.

To ensure that the shared LoRA matrices scale across multiple directions without favoring any single component excessively (which could result in parameter updates along only one direction and hinder task adaptability), we introduce a transformation component normalization matrix $\mathcal{G}$: 
\begin{equation} 
\mathcal{G}_i(x) = \frac{\exp(h(x)_i)}{\sum_{j=1}^{N} \exp(h(x)_j)} 
\end{equation} 
where $h(x)$ denotes the router layer matrix, calculated as $h(x) = W_rx$.

For a maximum of $N$ directions, the entire MoR module is expressed as: 
\begin{equation} 
\begin{split} 
o &= Wx + \Delta Wx\\
&\approx Wx + \sum_{i=1}^{N} \mathcal{G}_i(x) D_i x
\end{split}
\end{equation} 
where $N$ is the number of transformation spaces.

\subsection{Implementation in LLaMA}
In our implementation, we replace only the Feed-Forward Neural Network with the MoR plug-in. We denote the output of the FFN as $o$, and the forward pass during training can be expressed as:
\begin{equation}
o = Wx + \Delta Wx
\end{equation}
where $W \in \mathbb{R}^{d_{\text{out}} \times d_{\text{in}}}$ represents the pre-trained model parameters, and $\Delta W \in \mathbb{R}^{d_{\text{out}} \times d_{\text{in}}}$ represents the parameters updated during training.

The MoR plug-in is used to replace $\Delta W$, keeping all parameters of the base model frozen during training, and only the MoR parameters within the FFN are trained.

\begin{figure}[t]  
\centering  
\includegraphics[width=7.5cm]{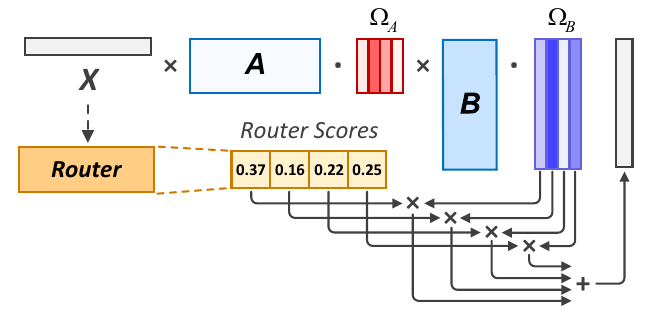}
\caption{Vectors stack method to speed up the training and inference process with matrix parallel computing through GPUs.}
\label{fig:impl}
\end{figure}

To ensure more efficient model training, we implement the MoR method in parallel on the Llama model, as shown in Figure~\ref{fig:impl}. For the trainable parameters $\lambda_A$ and $\lambda_B$, we stack them as matrices, which can be expressed as:
\begin{equation}
\begin{split}
\lambda_A \rightarrow \Omega_A , \hspace{0.2cm} \lambda_B \rightarrow \Omega_B
\end{split}
\end{equation}
where $\Omega_A \in \mathbb{R}^{N \times r}$, $\Omega_B \in \mathbb{R}^{N \times d_{\text{out}}}$, and $N$ is the number of specified expert hyperparameters.

During the forward computation, for $N$ Rank experts, the transformation vector can be matrixed using $\Omega_A$ and $\Omega_B$ to accelerate training. The forward computation of Llama, can be expressed as:
\begin{equation}
\small
\begin{aligned}
o &= Wx + \Delta Wx\\
  &= Wx + \frac{\alpha}{r} \sum_{i=1}^N \mathcal{G}_i(x) \Omega_B \cdot B(\Omega_A \cdot (A x))
\end{aligned}
\end{equation}

\section{Experiments}

\begin{table*}[!ht]
\setlength\tabcolsep{4pt}
\footnotesize
    \centering
    \begin{tabular*}{1.0\textwidth}{@{\extracolsep{\fill}}@{}l c cccccc cc cc c @{}}
    \toprule
    \multirow{2}{*}{\raisebox{-0.5\height}{\textbf{Method}}} & \multirow{2}{*}{\raisebox{-0.5\height}{\textbf{\# Params}}} & \multicolumn{6}{c}{\scriptsize\textbf{Commonsense \& Reading Comprehension}} & \multicolumn{2}{c}{\scriptsize\textbf{LM}} & \multicolumn{2}{c}{\scriptsize\textbf{World Knowledge}}  & \multirow{2}{*}{\raisebox{-0.5\height}{\textbf{Avg.}}} \\
    \cmidrule(lr){3-8} \cmidrule(lr){9-10} \cmidrule(lr){11-12}
    & & \scriptsize\textbf{SIQA} & \scriptsize\textbf{PIQA} & \scriptsize\textbf{WG} & \scriptsize\textbf{ARC-E } & \scriptsize\textbf{ARC-C} & \scriptsize\textbf{Hella.} & \scriptsize\textbf{OBQA} & \scriptsize\textbf{Lam.} & \scriptsize\textbf{MMLU} & \scriptsize\textbf{NQ} \\
    \midrule
    LLaMA2-7b & - & 37.77 & 53.81 & 50.91 & 50.62 & 34.58 & 26.68 & 35.60 & 73.30 & 39.52 & 28.61 & 43.14 \\
    \midrule
    Full SFT & 6.8B & 63.97 & 67.25 & 55.64 & 77.77 & 56.61 & 42.21 & 59.40 & 72.99& 45.50 & 27.62 & 56.90  \\
    \midrule
    LoRA (R8) & 11.6M & 62.44 & 56.75 & 50.51 & 61.90 & 46.44 & 26.87 & 42.40 & 72.33 & 42.79 & 26.70 & 48.91  \\
    LoRA (R16) & 23.2M & 52.71 & 56.15 & 49.72 & 58.55 & 47.11 & 28.99 & 42.40 & 73.06 & 45.98 & 28.82 &  48.34 \\
    DoRA (R8) & 12.3M & 52.05 & 56.58 & 50.43 & 57.50 & 46.10 & 27.84 & 44.80 & 72.95& 46.10 & 27.89 & 48.22  \\
    DoRA (R16) & 23.9M & 52.61 & 56.47 & 49.71 & 58.90 & 47.79 & 29.64 & 42.00 & 73.12 & 46.06 & \textbf{28.95} &  48.53 \\
    \midrule
    MoELoRA & 24.7M & \underline{63.43} & 61.53 & 52.09 & 72.55 & \underline{54.48} & \textbf{43.32} & \underline{52.40} & 72.09 & 47.16 & 25.20 &  \underline{54.43} \\
    \midrule
    MoR (E1R8) & 12.9M & 61.51 & \textbf{64.04} & 50.04 & 49.36 & 40.00 & 38.66 & 40.00 & 71.10 & 46.58 & 26.07 &  48.74 \\
    MoR (E2R8) & 14.3M & 61.51 & \underline{63.71} & \textbf{53.99} & 68.08 & 54.46 & 42.64 & 51.40 & \textbf{73.28} & \underline{47.73} & 26.26 & 54.31  \\
    MoR (E4R8) & 17.7M & 61.41 & 63.36 & 52.41 & \underline{73.37} & 53.22 & 38.43 & 51.80 & \underline{72.73} & 47.11 & \underline{27.26} & 54.11  \\
    MoR (E8R8) & 23.2M & \textbf{64.07} & 61.26 & \underline{52.64} & \textbf{76.90} & \textbf{56.95} & \underline{42.96} & \textbf{55.80} & 72.60 & \textbf{47.74} & 26.51 &  \textbf{55.74} \\
    \bottomrule
    \end{tabular*}

    
    \caption{Comprehensive assessment of model's fundamental capabilities with different fine-tuning methods. Specially, 'E$a$' and 'R$b$' denotes the expert number is $a$ and the shared LoRA matrix rank is $b$. Bold text and underlined text denote the best and second-best results. The abbreviated task name LM denotes Language Modeling, WG denotes Winogrande, Hella denotes Hellaswag, OBQA denotes OpenBookQA and Lam denotes Lambada.}
    \label{table:main-results1}

\vspace{-0.4cm}
\end{table*}


\begin{table}[!htbp]
\centering
\small{\resizebox{\columnwidth}{!}{%
\begin{tabular}{@{}llll@{}}
\toprule
\textbf{Method} & \textbf{\# Params} &  \textbf{Initialization}  \\ \midrule
LoRA         & $2dr$          & $A \sim\mathcal{N}, B = 0$  \\
DoRA         & $2dr+d$          & $A \sim\mathcal{N}, B = 0$  \\
MoELoRA      & $2Ndr+Nd$       & $A_i, R\sim\mathcal{N}, B_i = 0$ \\
MoR          & $2dr+N(2d+r)$            & $A, R \sim\mathcal{N}, B = 0$\\
\bottomrule
\end{tabular}
}}
\caption{Method~configurations included in our study.
Formulas for the number of trainable parameters in each configuration as a function of number of components $N$, hidden size $d$, and low-rank $r$.}
\label{tab:methods}
\end{table}

\subsection{Setup}
\paragraph{Datasets} 
To demonstrate the effectiveness of our method in multi-task learning, we train the model on a series of well-curated datasets, enabling it to excel in multiple downstream tasks simultaneously. Details of the datasets are provided in Appendix \ref{appendix_a}.

\paragraph{Training} 

Our experimental framework utilizes the implementation on the PEFT package~\cite{peft} and is executed on 8 NVIDIA A800 GPUs (80GB). Our MoR plugin is integrated into the LLaMA feedforward neural network, specifically within the `up\_proj', `down\_proj', and `gate\_proj' layers. Each layer is initialized with multiple sets of transformation vectors designed to capture diverse aspects of information across tasks. Meanwhile, task-shared information is stored in the shared LoRA mapping matrices, denoted as $A_s$ and $B_s$.


For the shared LoRA parameters, we set the hyperparameters $\alpha$ and $r$ to 32 and 8, respectively. Dropout rate is applied at 0.05, and the learning rate is set to $2 \times 10^{-4}$. During the model training phase, we freeze all base model parameters; only the transformation vectors in MoR, shared LoRA parameters, and direction constraint matrices are fine-tuned. To ensure fairness, the global batch size is consistently set to 8, and the input max length is set to 1024 across all experiments.

\paragraph{Evaluation} 
We employ the OpenCompass evaluation framework~\cite{2023opencompass} to conduct a thorough assessment of the model's performance, focusing on three primary tasks. The details about the evaluation are shown in Appendix \ref{appendix_b}.

\subsection{Baseline}



\paragraph{Supervised (Full) Fine-tuning}
We fine-tune all model parameters with the training set. Full parameter fine-tuning generally achieves the best performance in instruction-tuning tasks.

\paragraph{Low-Rank Adaptation}
We apply both vanilla LoRA and its improved version, DoRA~\cite{hu2021lora,liu2024dora}, for model fine-tuning. Specifically, We apply LoRA and DoRA to all linear layers in feed-forward network, and fine-tune the output, up, gate, and down projection weights in these layers. The scaling factor $\alpha$ and dropout rate are set to 32 and 0.05, respectively.

\paragraph{Multi-LoRA Tuning}
We also conduct experiments with a method that integrates multiple LoRA modules with the MoE mechanism, MoELoRA~\cite{liu2023moelora}. Specifically, we apply two LoRA experts with a rank of 8 and a router linear layer to all linear layers in the feed-forward network, fine-tuning the output, up, gate, and down projection weights. The LoRA scaling factor $\alpha$ and dropout rate are set to 32 and 0.05, respectively.


\subsection{Result}



\paragraph{Enhancement of Performance}
The results presented in Table~\ref{table:main-results1} demonstrate the effectiveness of our proposed MoR method, which simultaneously enhances various specialized capabilities by performing intrinsic transformations on LoRA. Specifically, MoR improves the average performance by 7.4\% over LoRA and by 7.21\% over DoRA. Furthermore, when compared with the MoELoRA method, which involves multiple LoRA ensembles, our method achieves a 1.31\% performance improvement over the simple ensemble approach. This improvement is attributed to the intrinsic transformation of LoRA and the use of scaling vectors to store specific knowledge while sharing the LoRA matrix. Our method achieves the best results on four of the six tasks in the Commonsense \& Reading Comprehension benchmark, and the OpenbookQA and MMLU tasks. However, performance decreases on Lambda tasks that were not included in the training data, and NQ task. This indicates that, while our approach greatly enhances learning capabilities, it also results in limited performance on out-of-distribution data.

\begin{table}[!t]
\small
  \centering
  \begin{spacing}{0.9}
    \setlength{\tabcolsep}{2.7mm}{
    \begin{tabular}{cccc}
    \toprule
    \textbf{\# Experts} & \textbf{\begin{tabular}[c]{@{}c@{}}\#R\textbf{ank}\end{tabular}} & \textbf{\begin{tabular}[c]{@{}c@{}}\textbf{\# Params}\end{tabular}} & \textbf{\begin{tabular}[c]{@{}c@{}}\textbf{Avg.}\end{tabular}} \\
    \midrule
    1         & 8         & 0.19\%             & 48.74        \\
    2         & 8         & 0.21\%             & 54.31       \\
    4         & 8         & 0.26\%             & 54.11        \\
    8         & 8         & 0.34\%             & 55.74        \\
    12         & 8         & 0.43\%             & 55.75        \\
    \midrule
    8         & 4         & 0.26\%             & 54.39        \\
    8         & 8        & 0.34\%             & 55.74       \\
    8         & 16       &  0.51\%             & 54.66        \\
    8         & 32       & 0.85\%             & 56.52       \\
    8         & 64       & 1.53\%             & 53.54       \\
    \bottomrule
\end{tabular} }%
\end{spacing}
  \caption{MoR performance varies based on the number of experts and LoRA rank across all benchmarks, including average results
  }
  \label{tab:sense-exp}
\end{table}

\begin{table}[t]
\centering
\small
\setlength{\tabcolsep}{4.5mm}{
\begin{tabular}{@{}lcc@{}}
\toprule
\textbf{Method} & \textbf{\#Params} & \textbf{Avg.}  \\ \midrule
Mean Pooling         & 17.4M          &51.39 \\
Balanced Router      & 24.7M       & 52.63 \\
Ours         & 24.7M          & 55.74  \\
\bottomrule
\end{tabular}
}
\caption{Average performance of MoR varies with different route  strategies.}
\label{tab:router}
\end{table}

\begin{figure}[t]  
\centering
\includegraphics[width=7.5cm]{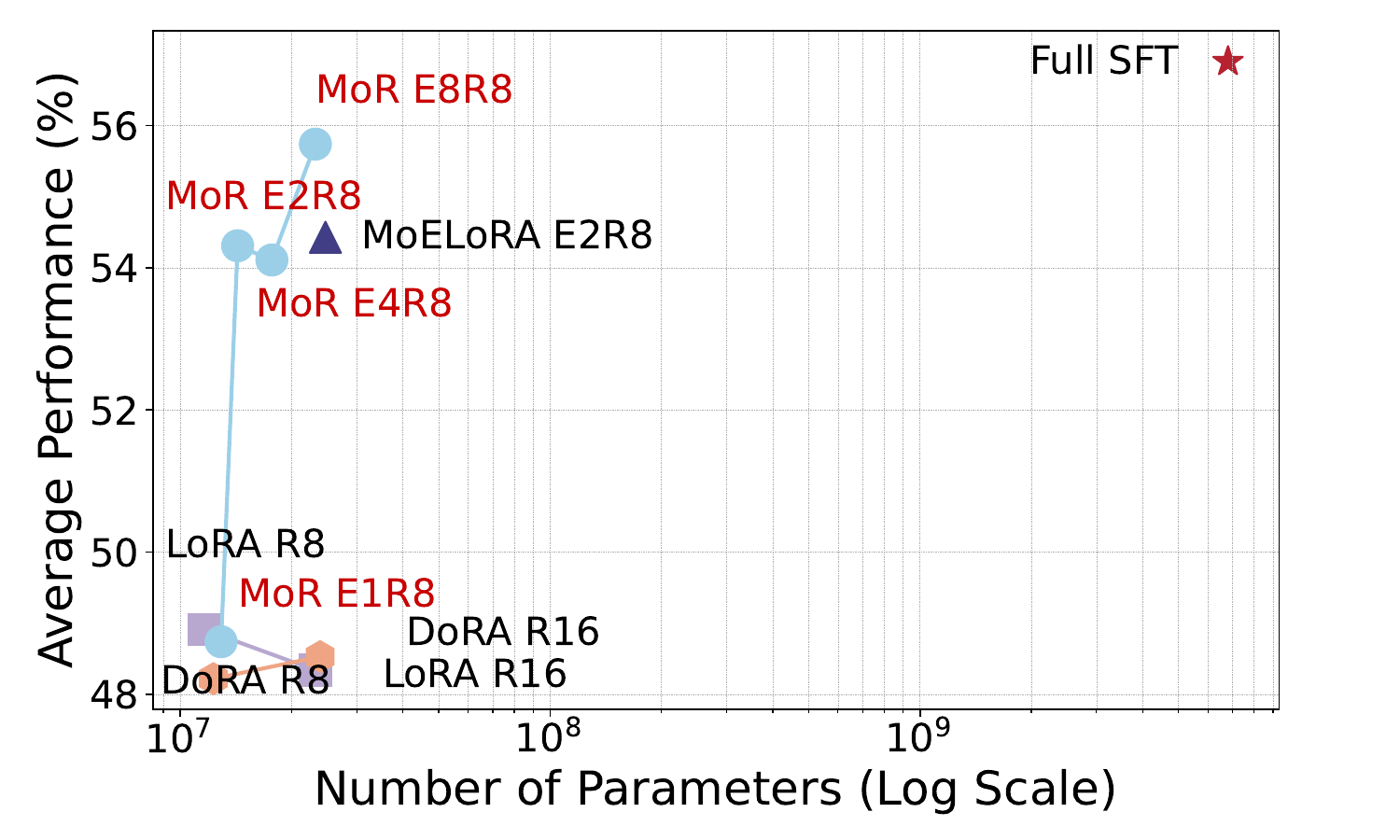}
\caption{Performance and parameter scale comparison between baselines and \textcolor[rgb]{0.74, 0, 0}{MoR}. \textcolor[rgb]{0.74, 0, 0}{MoR E1R8} indicates one expert with a shared LoRA matrix rank of 8.}
\label{fig:performance and parameter}
\end{figure}

\begin{figure}[t]  
\centering
\includegraphics[width=4cm]{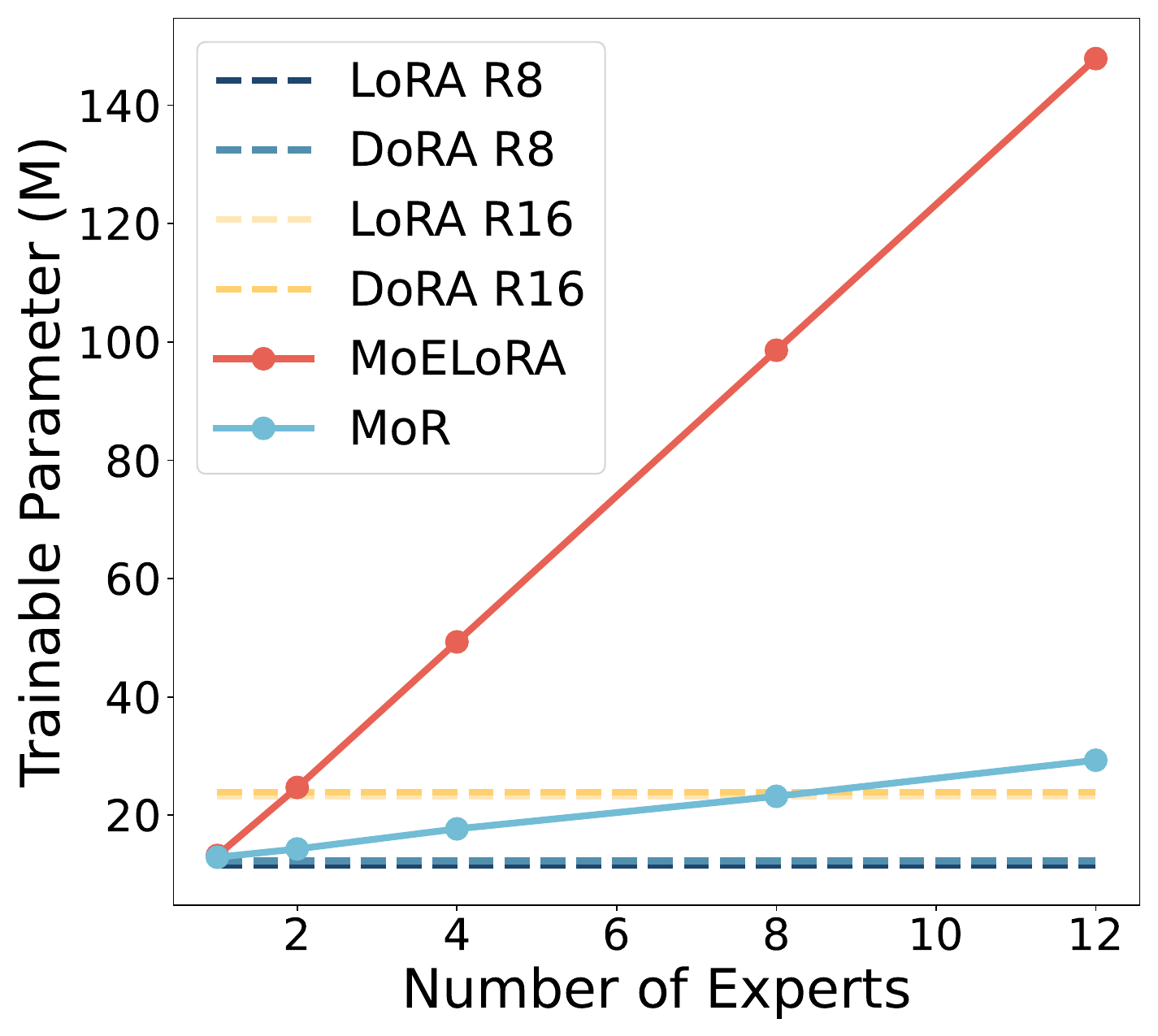}
\caption{Varies of nums trainable parameters on different experts nums.}
\label{fig:parameter}
\end{figure}

\paragraph{Parameter Efficiency}
Table~\ref{table:main-results1} and Figure~\ref{fig:performance and parameter} illustrate that when our model employs a single vector expert group, its performance matches that of LoRA and DoRA. This is because when the experts number comes to 1, our method is fundamentally equivalent to the DoRA approach. When increasing the number of vector expert groups to four leads to a substantial improvement in model performance without a significant increase in the number of parameters compared to LoRA and DoRA. Furthermore, when the number of vector expert groups is expanded to eight, our method surpasses the MoELoRA approach, which integrates multiple LoRA modules, by achieving a 1.31\% performance improvement while utilizing fewer parameters. Additionally, our method nearly approximates the performance of full parameter fine-tuning, attaining 98\% of its performance while using only 0.34\% of the parameters.
Moreover, as shown in Figure~\ref{fig:parameter} and Table~\ref{tab:methods}, our method demonstrates excellent expert scalability compared to MoELoRA. As the number of experts increases, the trainable parameters grow slowly compared to MoELoRA.

\subsection{Analysis}

\begin{figure}[t]
\centering
\subfigure[Average score with different experts numbers.]{
\includegraphics[width=3.65cm]{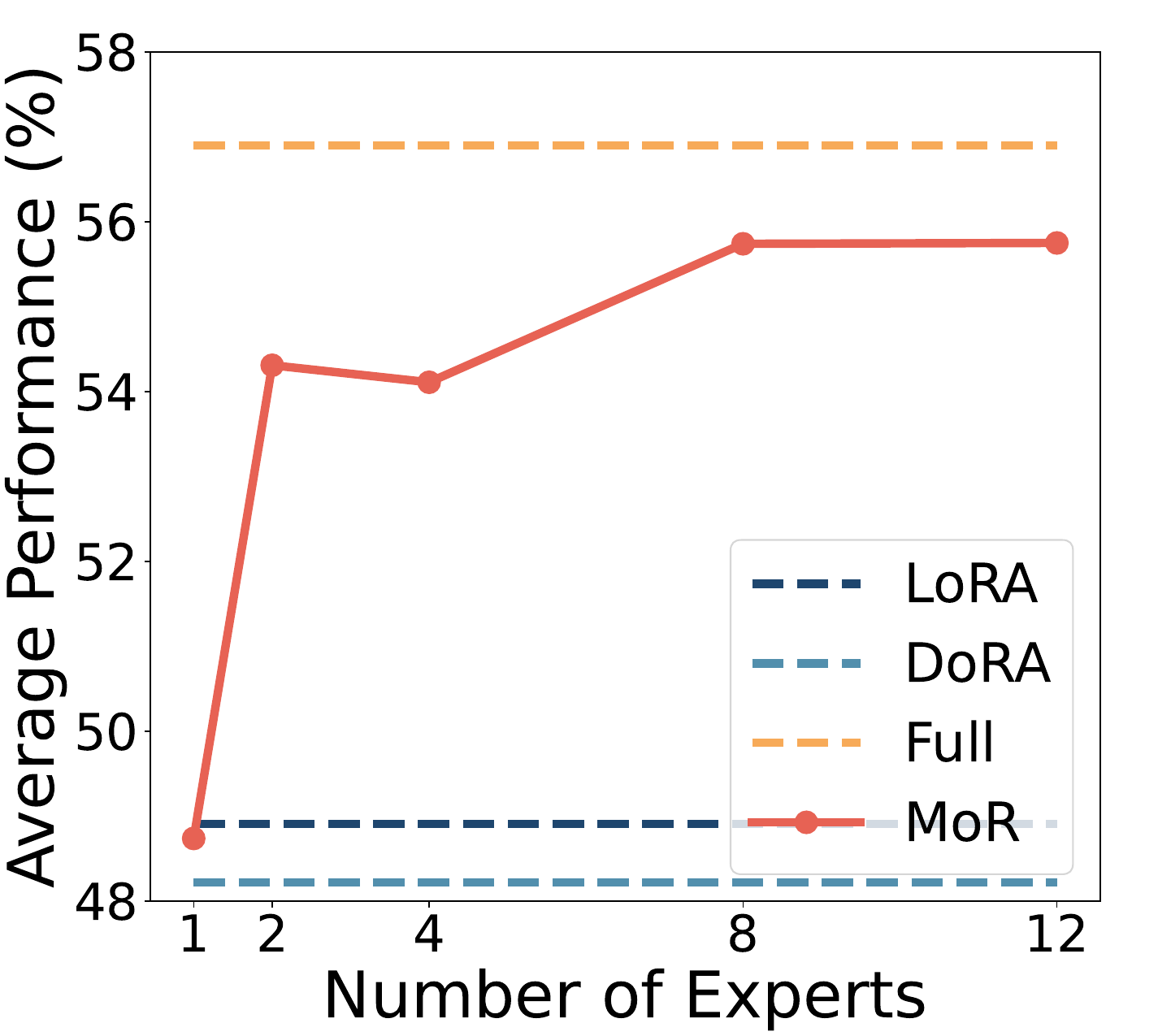}}
\hspace{0.05cm}
\subfigure[Average score with different LoRA rank.]{\includegraphics[width=3.65cm]{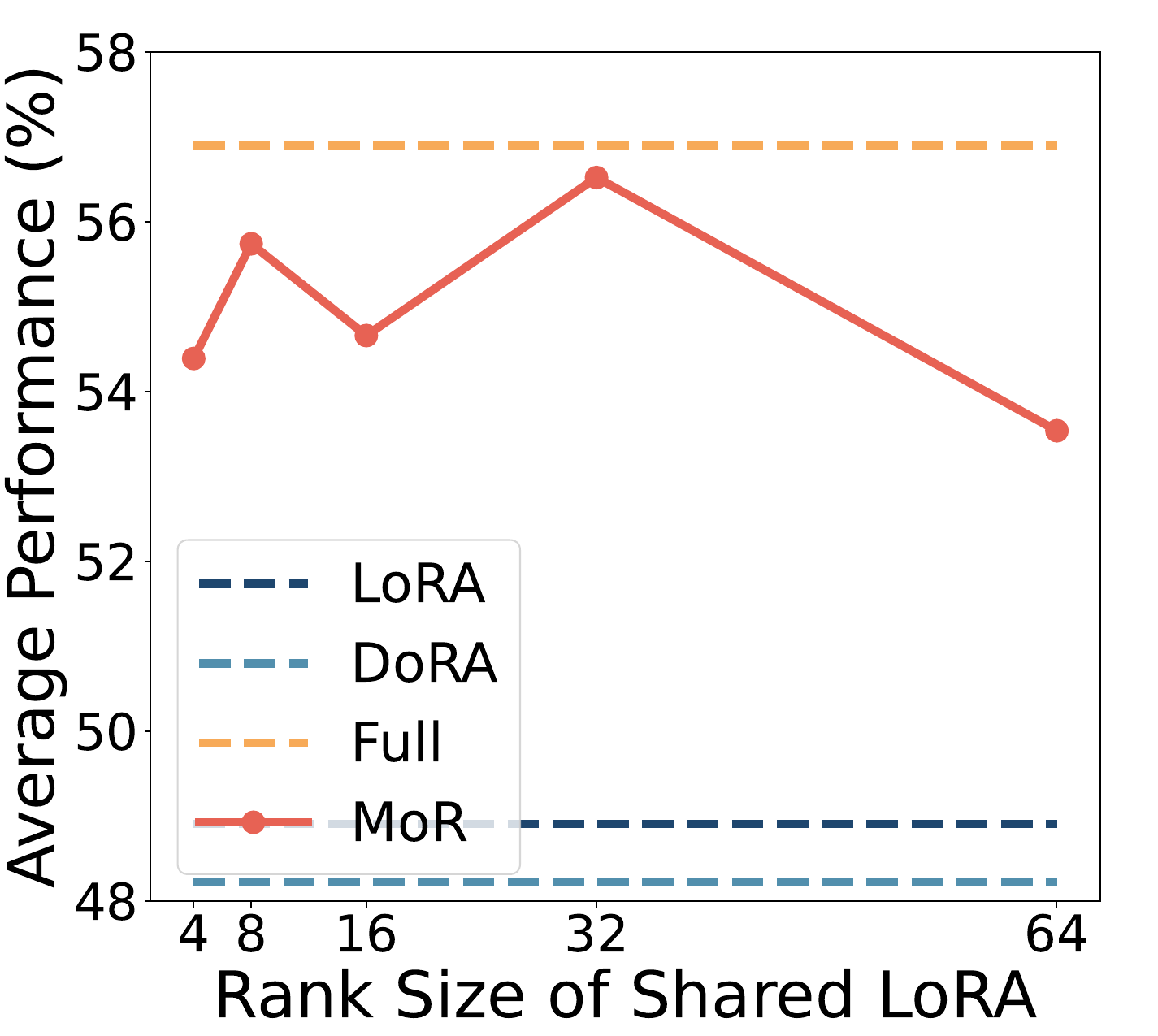}}
\caption{Average score with different experts nums and LoRA rank size. We use rank 8 as the LoRA and DoRA setting.}
\label{ablation study figures}
\end{figure}



\paragraph{Number of Vectors}
We conducted a series of analyses on the impact of the number of transformation vectors on model performance, including cases where the number of scaling vector experts is 1, 2, 4, 8, and 12, as shown in Table~\ref{tab:sense-exp}.


As is shown in Figure~\ref{ablation study figures}(a) When the number of experts is less than 8, the model's performance gradually improves as the number of experts increases, and reaches its peak at 8 experts. When the number of experts increases to 12, the model's performance decreases. This indicates that increasing the number of model experts can improve the performance within a certain range, but when it exceeds a certain number, the performance improvement will stagnate or even decrease.

\paragraph{Rank Size}
We also conducted a series of experimental analyses on the size of the shared rank, setting the rank of the shared matrix to 4, 8, 16, 32, and 64, as shown in the Table~\ref{tab:sense-exp}.

As shown in Figure~\ref{ablation study figures}(b), When the rank of shared LoRA is less than 32, the model's performance will be improved to a certain extent as the rank size increases. However, when the rank size exceeds 64, the model's performance will decrease to a certain extent. This finding is the same as the report of native LoRA~\cite{hu2021lora}. Continuously increasing the rank size of shared LoRA does not necessarily bring about performance improvement but will instead cause performance degradation to a certain extent.

\paragraph{Router Type} We also explored the role of Router in our method. We set up three comparative experiments: using the random learnable router (ours), directly using the Mean Pooling router, and constraining the router with an auxiliary balance loss.

The Mean Pooling router averages the router weighted output for each expert, and it's not learnable. The Balanced router takes the auxiliary router balance loss to constrain the router learning process, leading the router weight to have a balanced distribution. In the implementation, we adapt the auxiliary loss of switch-transformer into the experiment~\cite{fedus2022switch}.

As shown in Table~\ref{tab:router}, the routing method in our proposed method achieved the best results, improving by 4.35\% compared to the average weighted method and 3.11\% compared to the balanced router method. This means random learning of the router makes model's performance better than every expert having a similar routed weight affected by auxiliary loss or direct average.

\begin{figure}[!ht]
\centering
\includegraphics[width=6.5cm]{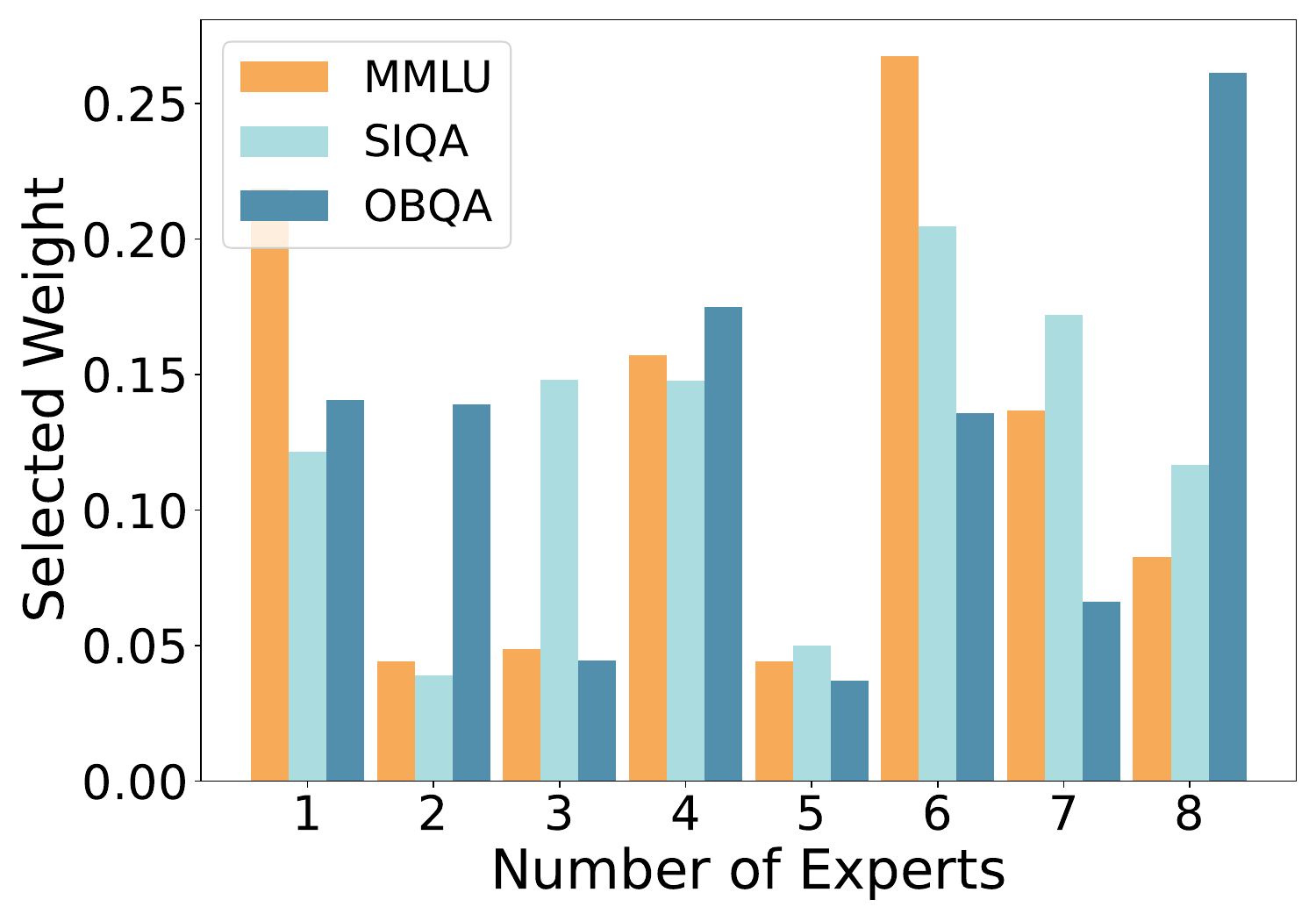}
\caption{Router weight distribution between different experts on different tasks, including MMLU, SIQA and OpenBookQA.}
\label{fig:router distribution}
\end{figure}

\paragraph{Router Analysis}
We also visualize the router output distribution to study the  influence of the learnable softmax router on different tasks. We choose 3 different datasets from commonsense and reading comprehension, language modeling and world knowledge benchmarks, including MMLU, SiQA, and OpenbookQA tasks. To visualize the effect of the router, We use the router of 31st layer's gate matrix of the Llama model as an example. 

As shown in Figure~\ref{fig:router distribution}, the MMLU data tends to utilize experts 1, 4, and 6, with these experts outputting significantly larger routing weight values compared to others. This indicates that different experts have different preferences when handling different tokens and they are task-specific experts.

\section{Related Work}
\paragraph{LoRA and its variants}
Low-Rank Adaptation~\cite{hu2021lora} is motivated by the discovery that a low intrinsic dimension within large parameters plays a crucial role in Large Language Models~\cite{aghajanyan2020intrinsic}. It introduces two trainable low-rank matrices into each layer and reduces storage and task-switching overhead by sharing pre-trained models across different tasks.


Subsequently, numerous methods have been proposed to enhance the performance and efficiency of LoRA and enlarge the learnable rank. Inspired by random projections, VeRA~\cite{kopiczko2023vera} shares two frozen random matrices across all layers and only trains a set of vectors for every VeRA component. Additionally, Tied-LoRA~\cite{renduchintala2023tied} enhances parameter efficiency by sharing trainable LoRA matrices across all layers and introducing the trainable vectors to improve the performance of LoRA. Moreover, DoRA~\cite{liu2024dora} enhances the model performance by decomposing LoRA into direction and amplitude with only one learnable vector added. However, these methods just transform a single group of LoRA in a single direction, resulting in poor learning ability of the model on multiple tasks. Our proposed method can more effectively learn multiple task data by performing intrinsic changes on the LoRA matrix at various angles and finally integrating them.

\paragraph{MoE-style PEFT}
The Mixture of Experts framework replaces the model with sparsely activated expert modules, boosting model capacity and throughput speed~\cite{jacobs1991adaptive}. Token-level MoE architectures are now widely used in pre-trained language and vision models\cite{shazeer2017sparsely,liu2024deepseek,wu2024mixture}.


To make the model better learn these multi-task datasets, MoELoRA~\cite{liu2023moelora} leverages LoRA as a plugin-type method for task-specific and multi-task tuning, particularly in medical applications. LoRAMoE~\cite{dou-etal-2024-loramoe} introduces an MoE-style plugin and a Localized Balancing Constraint to mitigate world knowledge forgetting in LLMs while enhancing the model’s multi-task learning abilities. Nevertheless, these methods exhibit significant parameter redundancy across different LoRAs. To achieve parameter efficiency, our proposed method substantially improves the model's multi-task learning capabilities by performing multi-dimensional intrinsic transformations on shared LoRA.

\section{Conclusion}

In this work, we first analyze the advantages and challenges associated with expanding the rank of LoRA and introduce a framework that equates the integration of multiple LoRAs to an expansion of LoRA rank. Building on this foundation, we propose the MoR framework. MoR learns LoRA transformations specific to different tasks based on varying inputs and efficiently integrates multi-rank information. By fully leveraging the foundational knowledge captured by LoRA, MoR reduces the complexity of learning and enhances multi-task capabilities. Compared to baseline methods, MoR achieves an absolute performance improvement of 1.31\% with only 93.93\% of the parameter count, establishing itself as a state-of-the-art LoRA approach that balances efficiency and performance.

\section*{Limitations}
In this section, we discuss the potential limitations
of our proposed method MoR. Firstly, although we have demonstrated our advantage while enhancing the downstream ability of the LLMs with multi-task datasets, we limit the model size to 7B due to resource and time constraints. Further work will be conducted on the larger LLMs to understand the influence of large-scale SFT on these LLMs and to boost their multi-task abilities. Secondly, Our method is similar to other MoE-style LoRA methods that can't merge the plugin parameters with a pre-trained model like vanilla LoRA, which will lead to latency during the inference period. Future work will be conducted on the rule-guided router to solve the problem. 

\bibliography{custom}

\appendix

\section{Train Data Details} \label{appendix_a}
Due to the computation constrain, We utilize the subset of dataset Tulu-v2~\cite{ivison2023camels} to conduct our experiment, which is composed with the subset of Flan-v2, Open Assistant, GPT-4 Alpaca, Code Alpaca, LIMA, WizardLM, Orca, Hardcoded and Science.

\section{Evaluation Details} \label{appendix_b}
We evaluate the different methods on three kinds of tasks, which are widely used in LLMs abilities assessment. The details about different tasks are illustrate below.
\paragraph{Commonsense \& Reading Comprehension Task}    These kinds of benchmarks are essential components in the evaluation of LLMs. Experiments on these tasks offer valuable insights into a model's ability to comprehend and interpret human language. Specifically, these tasks test a model's capacity to apply everyday knowledge, perform logical reasoning, and extract relevant information from text. In detail, We report 0-shot accuracy results for socialqa(SIQA.)~\cite{sap2019socialiqa}, PIQA~\cite{bisk2020piqa}, WinoGrande (Wino.)~\cite{sakaguchi2021winogrande}, ARC Easy(ARC-E.)\cite{clark2018think}, HellaSwag (Hella.)~\cite{zellers2019hellaswag}, and ARC Challenge (ARC-C.)\cite{clark2018think}.

\paragraph{Language Modeling Task} These kinds of benchmarks assess a model's proficiency in understanding and generating human language. This task evaluates the model's ability to leverage domain-specific knowledge, engage in complex reasoning, and maintain coherence across extended passages. By measuring a model’s effectiveness in capturing intricate linguistic patterns and understanding context, this task provides guidance for advancing sophisticated language modeling techniques. In detal, We report 0-shot accuracy results for OpenBookQA(OBQA.)~\cite{mihaylov2018can} and Lambada(Lam.)~\cite{harinilong}.

\paragraph{World Knowledge Task} These kinds of benchmarks are critical for evaluating a model's comprehension of real-world information. This task assesses the model’s ability to retrieve, synthesize, and respond accurately to diverse queries. It's particularly useful in identifying catastrophic forgetting in LLMs, as it measures how well a model retains broad, multi-domain knowledge over time. In detail, We report 0-shot accuracy results for MMLU~\cite{hendrycks2020measuring} and Natural Question(NQ.)~\cite{kwiatkowski2019natural}.

\end{document}